\documentclass[conference]{IEEEtran}
\IEEEoverridecommandlockouts
\usepackage{cite}
\usepackage{amsmath,amssymb,amsfonts}
\usepackage{algorithmic}
\usepackage{graphicx}
\usepackage{textcomp}
\usepackage{xcolor}
\usepackage{comment}
\usepackage[numbers]{natbib}
\usepackage{enumitem}

\def\BibTeX{{\rm B\kern-.05em{\sc i\kern-.025em b}\kern-.08em
    T\kern-.1667em\lower.7ex\hbox{E}\kern-.125emX}}
\begin{document}
\title{Language Modeling through Long Term Memory Network\\
{\footnotesize \textsuperscript{*}The article was accepted to IJCNN 2019}
\thanks{© 2019 IEEE. Personal use of this material is permitted. Permission from IEEE must be obtained for all other uses, in any current or future media, including reprinting/republishing this material for advertising or promotional purposes, creating new collective works, for resale or redistribution to servers or lists, or reuse of any copyrighted component of this work in other works. 
}
}

\author{\IEEEauthorblockN{Anupiya Nugaliyadde}
\IEEEauthorblockA{\textit{College of Science, Health, Engineering and Education} \\
\textit{Murdoch University}\\
Murdoch, Australia \\
a.nugaliyadde@murdoch.edu.au}
\and
\IEEEauthorblockN{Kok Wai Wong}
\IEEEauthorblockA{\textit{College of Science, Health, Engineering and Education} \\
\textit{Murdoch University}\\
Murdoch, Australia \\
k.wong@murdoch.edu.au}
\and
\IEEEauthorblockN{Ferdous Sohel}
\IEEEauthorblockA{\textit{College of Science, Health, Engineering and Education} \\
\textit{Murdoch University}\\
Murdoch, Australia \\
f.sohel@murdoch.edu.au}
\and
\IEEEauthorblockN{Hong Xie}
\IEEEauthorblockA{\textit{College of Science, Health, Engineering and Education} \\
\textit{Murdoch University}\\
Murdoch, Australia \\
h.xie@murdoch.edu.au}

}
\maketitle
\begin{abstract}
Recurrent Neural Networks (RNN), Long Short-Term Memory Networks (LSTM), and Memory Networks which contain memory are popularly used to learn patterns in sequential data.  Sequential data has long sequences that hold relationships. RNN can handle long sequences but suffers from the vanishing and exploding gradient problems. While LSTM and other memory networks address this problem, they are not capable of handling long sequences (50 or more data points long sequence patterns). Language modelling requiring learning from longer sequences are affected by the need for more information in memory. This paper introduces Long Term Memory network (LTM), which can tackle the exploding and vanishing gradient problems and handles long sequences without forgetting. LTM is designed to scale data in the memory and gives a higher weight to the input in the sequence. LTM avoid overfitting by scaling the cell state after achieving the optimal results. The LTM is tested on Penn treebank dataset, and Text8 dataset and LTM achieves test perplexities of 83 and 82 respectively. 650 LTM cells achieved a test perplexity of 67 for Penn treebank, and 600 cells achieved a test perplexity of 77 for Text8. LTM achieves state of the art results by only using ten hidden LTM cells for both datasets.
\end{abstract}


\section{Introduction}
Natural language understanding requires processing sequential data. Natural language is time-dependent, and past information can influence the current and future output. Therefore, models which are capable of processing sequential data are required. Memory determines the models’ capability of recalling from past information. Sequential deep learning models have shown to achieve state-of-the-art results in natural languages understanding tasks such as question answering \cite{nugaliyadde2017reinforced}, machine translation \cite{bahdanau2014neural}\cite{yang2017multi}, and language modelling \cite{mikolov2014learning}\cite{ororbia2017learning}\cite{singh2017temporal}.

The memory networks have a recurrent behaviour which use outputs to influence the current output \cite{ororbia2017learning}\cite{hochreiter1997long}\cite{json2014memnet}. With the increase in sequence length, the effect on the current input is reduced, and after a certain number of steps the effect on the current input becomes invisible. In order to understand a language, the model is required to learn from past knowledge. Relevant information to understand language is spread throughout the sequence. Therefore, long-term memory is required for natural language understanding \cite{mikolov2014learning}\cite{ororbia2017learning}.

Recurrent Neural Network (RNN)s are capable of handling long sequences but suffer from the exploding and vanishing gradient descent \cite{bengio1994learning}\cite{mikolov2014learning}. In order to overcome the issue Long Short-Term Memory Networks (LSTM)s \cite{hochreiter1997long}, Simple Recurrent Network \cite{mikolov2014learning} and Memory Network \cite{json2014memnet} clip the gradient. These models still suffer from the problem of vanishing gradient when the sequences are long. The gradients of non-linear functions are close to zero, and the gradient is back propagated through time while multiplied. When the eigenvalues are small, the gradient will converge to zero rapidly. Therefore, these models are capable of only handling short-term dependencies.

LSTM, GRU and SRN proposed by Mikolov \cite{mikolov2014learning} use gates to control the vanishing gradient problem. These gates control the vanishing gradient problem. The gates control the memory sequence and prevents the overflow of data. The forget gate in the LSTM is a crucial element which forgets the past sequence \cite{gers1999learning}. The gates control or forget the previous sequences which influence the current input. Therefore, these memory networks do not handle long term sequences.

 Holding longer sequences in memory is important in properly understanding a language and it is also necessary in many long term dependency tasks\cite{nugaliyadde2019enhancing}. In order to remember long sequences as well as to prevent the learning model from suffering from the vanishing gradient problem, Long Term Memory Network (LTM) is introduced in this paper. LTM does not forget the past sequences. LTM incorporates the past outputs and current inputs. LTM generalises the past sequences and gives a higher emphasis on the new inputs in order to support natural language understanding. LTM was tested for long-term memory dependency based language modelling tasks. LTM is tested on Penn Treebank and Text8 datasets and it outperformed the current state-of-the-art memory network models.
\section{Background}
Long term memory dependencies require learning from patterns. Memory networks are used in order to learn long term dependencies \cite{nugaliyadde2017reinforced}. Memory networks including RNN and LSTM are used for many natural language tasks such as question answering, speech to text, language modelling and time series analysis \cite{nugaliyadde2017reinforced}\cite{graves2014neural}\cite{sukhbaatar2015end}\cite{weston2015towards}\cite{boukoros2017modeling}.These memory networks have shown to achieve state-of-the-art results in benchmark datasets. However, RNN, LSTM and other memory networks perform differently from each other, and each has its own merits.

RNN is capable of handling infinite continuous sequence \cite{pascanu1986learning}\cite{salehinejad2016learning}. It takes an input and passes the value continuously. The output is looped back and combined with the input \cite{pascanu2013difficulty}. The long term dependencies learning fails due to exploding and vanishing gradient problem \cite{hochreiter2001gradient}. This is due to the direct influence of the past information to the current input $x_{t} $ \eqref{eq1}. The internal state $S_{t}$ for a current input of RNN can be defined as:
\begin{equation}S = activation(U_{x_{t}} + W_{s_{t-1}}) \label{eq1} \end{equation} 
where activation can be any activation function (e.g. tanh, Relu), $U$ the weight for the current input, W being the weight for the past input state $S_{t-1} $ \cite{pascanu1986learning}. Therefore, the overall output would be affected by the past outputs. When $W_{s_{t-1}}$ is added to the weight of the current input $U_{x_{t}}$, the past state $S_{t-1}$ directly affects current state $S_t$ as shown in \eqref{eq1}.

LSTM was introduced in order to handle the vanishing and exploding gradient problem \cite{hochreiter1997long}. The forget gate was later added to the original LSTM. This is capable of preventing the internal state from growing indefinitely and handling the network break \cite{gers1999learning}. The forget gate resets the cell state when the it decides on forgetting the past sequence. The cell state holds the past inputs with the network or resets the cell state to forget past information held in the network. LSTM has shown to be a stable model that is not affected by the vanishing and exploding gradient problem \cite{hochreiter2001gradient}. However, the LSTM is only capable of handling short-term dependencies \cite{salehinejad2016learning}. 

Traditional memory networks (RNN and LSTM) have shown to handle natural language understanding tasks \cite{young2018recent}. RNN is capable of handling continuous data streams which are entered into the network as in speech recognition \cite{chen1998rnn} and language modelling \cite{mikolov2014learning}\cite{mikolov2010recurrent}\cite{singh2017temporal}. LSTM has shown to perform more complex tasks such as question answering \cite{nugaliyadde2017reinforced}\cite{weston2015towards}. The traditional memory networks and specified memory networks (Dynamic Memory Network \cite{kumar2016ask} and Reinforced Memory Network \cite{nugaliyadde2017reinforced}) benefit learning from longer dependencies in order to understand language. Longer dependencies are captured by adding more hidden layers. The hidden layers would also contribute towards the vanishing and exploding gradient. Therefore, forgetting the past sequences is one main approach used in memory networks \cite{ororbia2017learning}\cite{gers1999learning}. This affects on long-term dependency. 

The vanishing and exploding gradient is one of the most problematic issues in memory networks through backpropagation \cite{lecun2015deep}. Memory network trained by deriving the gradients of the network weights using backpropagation and chain rule. Consider a long sequence which has more than 30 words as the input,  ``\textit{I was born in France. I moved to UK when I was 5 years old ... I speak fluent French}". Using language models the last word of the paragraph ``\textit{French}" requires learning through a long dependency from the first word France. Passing the paragraph through an RNN can cause the vanishing and exploding gradient problem \cite{pascanu2013difficulty}. This problem occurs while the RNN is training. Gradients from the deeper layers have to go through matrix multiplications using Chain Rule, and if the previous layers have small values, it declines exponentially \cite{pascanu2013difficulty}. These gradient values are insignificant to the model to learn from; this is vanishing gradient problem. If the gradient is large, it gets larger and explodes which negatively affects the model’s training; this is the exploding gradient problem.

Clipping the gradients which places a predefined threshold value which changes the gradient length and attempts to control the vanishing and exploding gradient problem of RNN \cite{pascanu2013difficulty}. Gradient clipping affects the convergence of the gradient. LSTM and other memory networks avoid vanishing and exploding gradient by using gates which controls the passing the past outputs to the current input \cite{sak2014long}. Clipping also requires a target to be defined at every time step which increases the complexity \cite{pascanu2013difficulty}.  Memory networks including LSTM forget the past outputs which the network deems irrelevant. Attention-based memory networks \cite{chorowski2015attention} avoid vanishing and exploding gradient by focusing on only a few factors which are relevant to the tasks. These methods used to avoid the vanishing and exploding gradient prevents prolonging the memory of the network. According to the example, either the sequence is long, or the model does not identify relevancy in ``\textit{France}", it is removed from the memory. Model not knowing ``\textit{France}" would directly influence the model in predicting the last word ``\textit{French}".

Long-term memory network should have the capability of holding all the past sequences and not be affected by the vanishing or exploding gradient.

\section{Proposed Methodology for Long Term Memory}
\begin{figure}[htbp]
\centerline{\includegraphics[scale=0.45] {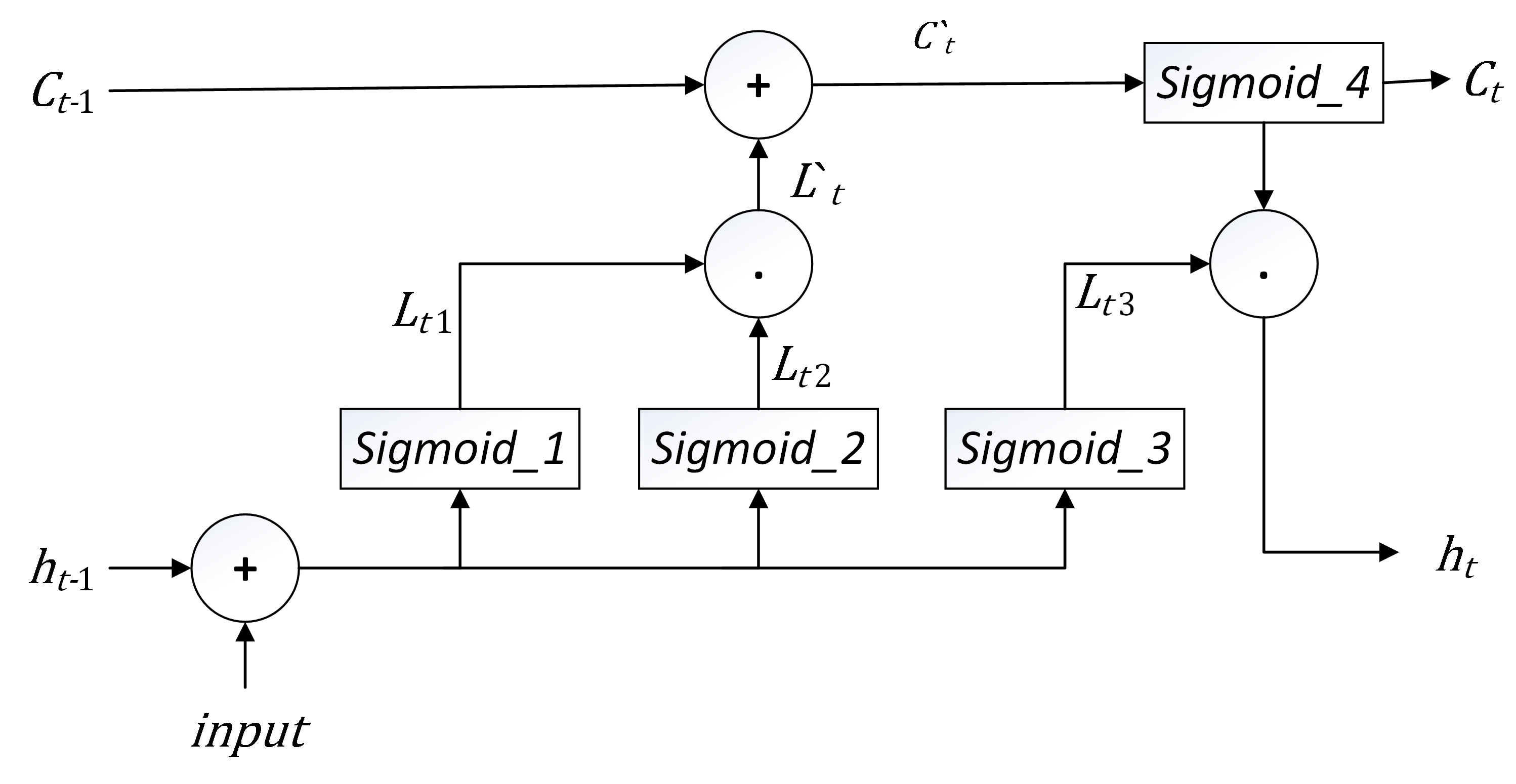}}
\caption{Long-Term Memory cell, the arrows show the data flow from within the cell. . indicates the dot product between the two vectors and + indicates the sum of the two vectors.}
\label{fig}
\end{figure}

The proposed model has two main objectives: 1) to handle longer sequences; and 2) to overcome the vanishing gradient. The proposed LTM is structured such that it is capable of holding and generalizing old sequences (Fig.~\ref{fig}) and give an emphasis on the recent information. Fig.~\ref{fig}. shows a single cell LTM which holds long-term memory which generalises the past sequences.

 Retaining longer memory sequences is a crucial requirement in natural language understanding since the past sequences affect the current inputs \cite{cambria2014jumping}. Furthermore, LTM gives an emphasis weight to the current input. The LTM holds three states:
\begin{enumerate}[label=(\alph*)]
\item input state: handles the current input to pass on to the output
\item cell state: carries the past information through each step to the other step.
\item output state: handles the current output and passes the output to the cell state.

\end{enumerate}
The LTM’s functionality relies on the gate structure within it. The LTM cell contains four gates with the first three gates impact on the inputs and the last gate controlling and generalising the cell state. However, the LTM’s cell state does not reset itself similar to LSTM’s forget gates function \cite{gers1999learning}. Therefore, LTM is capable of holding longer sequences in memory. The following sections provide the detail of the architecture.

\subsection{Input state}
The input is combined with the previous output and passed on to the Sigmoid\_1 as shown in  \eqref{eq2}. Equation  \eqref{eq2}, \( \sigma\)  indicates the sigmoid functions and $W_{1}$ is the weight for the gate. The $L_{t1}$ is the by-product which generates an effect on the LTM cell which depends on the current input and the previous output. 
\begin{equation}
L_{t1} = \sigma (W_1(h_{t-1}+input))
\label{eq2}
\end{equation}

Similarly, \eqref{eq3} shows a similar functionality with different weight $W_2$ which gives a higher impact on the current input although scaled through the sigmoid functions (Sigmoid\_1 and Sigmoid\_2). These two equations \eqref{eq2} and \eqref{eq3} support long-term memory by emphasising on the current input and adds on to the past input. $W_2$ is the weight for the gate represented by \eqref{eq3}.
\begin{equation}
L_{t2} = \sigma (W_2(h_{t-1}+input))
\label{eq3}
\end{equation}

In order to emphasise the current input to effect on the output $L_{t1}$ and $L_{t2}$ are passed through a dot operation to create $L'_t$. $L'_t$ is created as showin in \eqref{eq4}.
\begin{equation}
L'_{t} = L_{t1} . L_{t2}
\label{eq4}
\end{equation}

$L'_{t}$ amplifies the effect of the current input and past output. $L'_{t}$ is passed on to the cell state, which would be carried along to the future sequences.  $L'_{t}$ amplifies the current inputs’ effect on the output. 

\subsection{Cell state}
Cell state similar to LSTM's cell state \cite{hochreiter1997long} carries forward the past outputs to the present cell. Natural language understanding requires both past output and current inputs. The current input is emphasised over the past outputs. Therefore, $L'_t$ has a higher value combining the current input which is passed on to the cell state as shown in \eqref{eq5}. Therefore, the output would have a higher effect on the current input. As shown in \eqref{eq5}, $C'_t$, the current cell state combines the current input $L'_t$ and the past output $C_{t-1}$.
\begin{equation}
C'_{t} = L'_{t} + C_{t-1}
\label{eq5}
\end{equation} 
The final cell state $C_t$ as shown in \eqref{eq6} is calculated using the $C'_t$ and passing through the Sigmoid\_4. Through this, the LTM scales the cell state $C_t$. The cell state carries on a scaled value to the final output state. $W_4$ is the weight for the \eqref{eq6}.
\begin{equation}
C_{t} = \sigma (W_4 C'_t)
\label{eq6}
\end{equation}

\subsection{Output state}
Equation \eqref{eq7} shows the direct influence on the output of a given LTM cell. $L_{t3}$ directly influences the output by passing the current input. $W_3$ is the weight for the  \eqref{eq7}. 
\begin{equation}
L_{t3} = \sigma (W_3(h_{t-1}+input))
\label{eq7}
\end{equation}
The cell state $C_t$ and the $L_{t3}$ are joined together and combined through the dot operation. The $C_t$ and the $L_{t3}$ create the final output $h_t$. Equation \eqref{eq8} shows the final output creation. $h_t$ has a higher impact through the current input as well as the past outputs. Therefore, the impact from both the past and the current input are combined as shown in \eqref{eq8}.

\begin{equation}
h_{t} = C_t . L_{t3}
\label{eq8}
\end{equation}

The output $h_t$ and $C_t$, is passed on to the next time step, which is shown in (Fig.~\ref{fig2}). LTM is used as a cell, and the cell passes the $C_t$ and $h_t$. This also shows how the cells passes the past outputs on and combine with the current inputs. 

\begin{figure}[htbp]
\centerline{\includegraphics[scale=0.45] {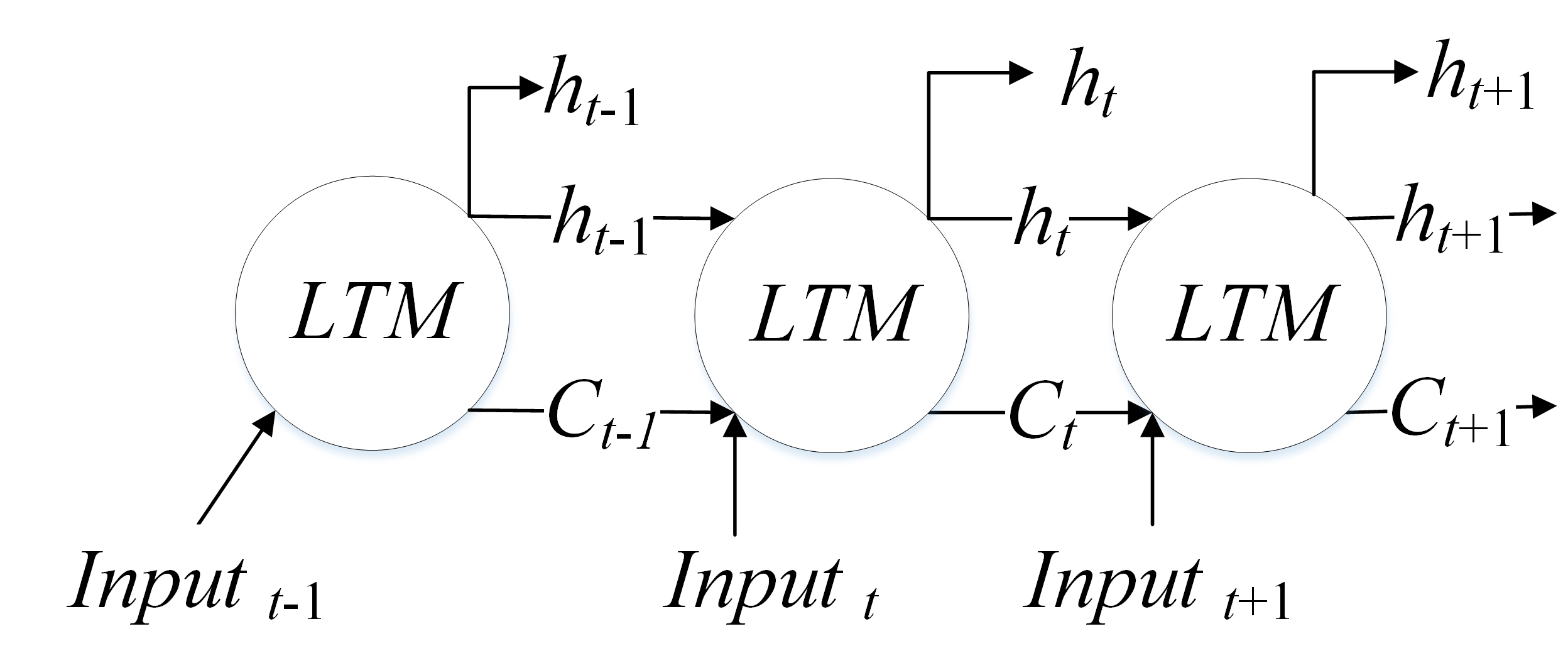}}
\caption{Long-Term Memory cells connected. The figure also illustrates the data passed on in the cell state and how the output is passed on from one LTM cell to another.}
\label{fig2}
\end{figure}

\section{Experimentation}
In order to demonstrate the long-term dependency learning, LTM is tested on language modelling. Three types of experiments are conducted to evaluate the LTM using Penn treebank dataset and Text8 dataset. Penn treebank dataset contains 2499 stories of Wall Street Journal. These stories are in raw text format. Text8 dataset contains over 240000 Wikipedia articles. Articles from both datasets contain long relationship dependencies between words. LTM is evaluated on the two datasets against the current state of the art models, and finally, LTM is evaluated against itself by changing the number of cells to find the best cell size which generates the best results.

LTM was first evaluated on Pennbank dataset \cite{taylor2003penn}. Similar to Mikolov et al. model \cite{mikolov2011extensions}, it consists of pre-processing the data and the training size of 930K tokens, validating the size of 74K tokens and testing size of 82K tokens. The dataset has a vocabulary of 10K words. In order to match with the current state-of-the-art model experiments, 300 LTM cells are used.

Second dataset Text8 \cite{mikolov2014learning} has 44K vocabulary from Wikipedia. The dataset has 15.3m training tokens, 848K validation tokens and 855K test tokens. The settings are similar to \cite{xie2017data}. Words which occur ten times or lower are placed as an unknown token. 500 LTM cells are used in the experiments. 

In order to evaluate the model on its performance, the cell number is gradually increased and tested for both Pennbank dataset and Text8. The experiment conditions are the same as the above experiments except for the number of layers. All the learning models on the Penn Treebank dataset follow similar \cite{mikolov2011extensions} and experiments on Text8 follows \cite{mikolov2014learning} this includes the inputs with the hyper-parameters.

\section{Results}
LTM’s long term memory is tested on Penn Treebank dataset and Text8 dataset. The results are validated using perplexity shown in \eqref{eq9}. Perplexity is the inverse probability of the test set, normalized by the number of words. The lower the perplexity the better the model.
\begin{equation}
Perplexity(W) = \sqrt[N]{\prod _{i=1}^{N}\frac{1}{P(w_i|w{_1}...w_{i-1})}}
\label{eq9}
\end{equation}

The first experiment was based on the Penn treebank dataset. Results are shown in Table I. LTM is tested against the traditional memory and recurrent networks and the current state of the art models (Delta-RNN). RNN which had the lowest performance over the tested models with 300 hidden layers achieved a test perplexity of 129. This demonstrates that RNN is not capable of handling long-term dependencies. Although LSTM has outperformed the RNN, ultra-specific models which handle long-term memory outperforms the generalised models on long-term memory. LTM achieves a test perplexity 83 with 300 units, which is 20 points above the current state of the art results. Furthermore, LTM achieves the state of the art results at ten hidden layers (Table III).

LTM was also tested with the Text8 dataset with 500 hidden layers. The LTM was compared against the traditional memory networks and the current state of the art models (MemNet) (Table II). LTM has outperformed all the state of the art model by only using ten hidden units (Table III). The ultra-specified long-term dependency based memory networks have shown to outperform the generic memory networks.

LTM was tested on Text8 and Penn treebank dataset by increasing its hidden layers in order to identify the best performing number of hidden layers. Table III shows validation and testing perplexity for Text8 and Penn Treebank while increasing the hidden layers. Table III also shows that LTM achieves the state of the art results with only ten hidden layers, in which other networks require 300 hidden layers or more to achieve state of the art results. The results are further improved by increasing the number of hidden layers. LTM achieved the best results for the Penn treebank dataset with 650 hidden layers. Furthermore, LTM achieved its best results for Text8 with 600 hidden layers.

\begin{table}[htbp]
\caption{Penn Treebank Validate and Test Perplexity. Perpl = Perplexity}
\begin{center}
\begin{tabular}{|c|c|c|c|}
\hline
\textbf{Models}&\multicolumn{3}{|c|}{\textbf{Penn Treebank}} \\
\cline{2-4} 
\textbf{} & \textbf{\textit{\# hidden layers }}& \textbf{\textit{Validate Perpl.}}& \textbf{\textit{Test Perpl.}} \\
\hline
RNN& 300& 133&129  \\
\hline
LSTM\cite{mikolov2014learning}& 300& 123&119  \\
\hline

SCRN\cite{mikolov2014learning}& 300&120 &115  \\
\hline
Delta-RNN & 300&- &102.7  \\
\hline
LTM& 300&85 &83  \\
\hline
\end{tabular}
\label{tab1}
\end{center}
\end{table}

\begin{table}[htbp]
\caption{TEXT8 VALIDATE AND TEST PERPLEXITY.  Perpl = Perplexity}
\begin{center}
\begin{tabular}{|c|c|c|c|}
\hline
\textbf{Models}&\multicolumn{3}{|c|}{\textbf{Text8}} \\
\cline{2-4} 
\textbf{} & \textbf{\textit{\# hidden layers }}& \textbf{\textit{Validate Perpl.}}& \textbf{\textit{Test Perpl.}} \\
\hline
RNN& 500& - &184  \\

\hline
LSTM\cite{mikolov2014learning}& 500& 122&154  \\
\hline
SCRN\cite{mikolov2014learning}& 500&161 &161  \\
\hline
MemNet\cite{sukhbaatar2015end}& 500& 118 &147  \\
\hline
LTM& 500&85 &82  \\
\hline

\end{tabular}
\label{tab1}
\end{center}
\end{table}

\begin{table}[htbp]
\caption{INCREASING THE NUMBER OF HIDDEN LAYERS THE VALIDATION AND TEST PERPLEXITY FOR TEXT 8 AND PENN TREEBANK DATASETS. Perpl = Perplexity}
\begin{center}
\begin{tabular}{|c|c|c|c|c|}
\hline
\textbf{\# hidden}&\multicolumn{2}{|c|}{\textbf{Text8}} &\multicolumn{2}{|c|}{\textbf{Penn Treebank}}\\
\cline{2-5} 
\textbf{layers} & \textbf{\textit{Train  Perpl.}}& \textbf{\textit{Test  Perpl.}}& \textbf{\textit{Train  Perpl.}} & \textbf{\textit{Test Perpl.}}\\
\hline
10& 103& 100 &100&99 \\
\hline
50& 101& 99 &98&97 \\
\hline
100& 99& 98 &95&92 \\
\hline
150& 97& 95 &90&89 \\
\hline
200& 95& 93 &88&86 \\
\hline
250& 93& 90 &87&85 \\
\hline
300& 90& 89 &85&83 \\
\hline
350& 89& 87 &82&80 \\
\hline
400& 87& 86 &79&78 \\
\hline
450& 86& 84 &77&76 \\
\hline
500& 85& 82 &74&72 \\
\hline
550& 81& 80 &72&70 \\
\hline
600& 79& 77 &69&67 \\
\hline
650& 79& 77 &68&67 \\
\hline
700& 79& 77 &68&67 \\
\hline

\end{tabular}
\label{tab1}
\end{center}
\end{table}

\section{Discussion}
The structure of the LTM, as shown in Fig.1. is designed in order to hold the inputs passed through the LTM cell and scale the output. The use of the sigmoid functions is a crucial aspect of maintaining a scaled output. Equation 6 is used to create the cell state and the output. The use of the sigmoid function in equation 6 scales the cell state in order to prevent exploding or vanishing gradient problem. Since the cell state is scaled and passed on from one-time stamp to the other time stamp the cell state value would not explode or vanish preventing the vanishing or exploding gradient. Vanishing and exploding gradient is the main reason for a memory network to forget or underperform. In order to prevent exploding or vanishing, gradient LSTM introduced the forget gate [7]. Using the forget gate the LSTM can handle longer sequences and forget the sequence when irreverent sequences are presented to the LSTM. However, the past sequences although not substantially relevant have an effect in long-term natural language understanding tasks. LSTM has a downfall in long-term memory. LTM scales the outputs and holds it in the memory. Therefore, even the long dependencies would affect the final output of the LTM.

LTM gives a high impact on the new inputs \eqref{eq4}. LTM combines $L_{t1}$ and $L_{t2}$ in order to pass a higher impact from the current input to the output as shown in \eqref{eq4}. Therefore, the LTM gives a higher priority to the new inputs, which is more relevant to the current output. Equation 8 shows the effect on the final output which combines both the processed input and the cell state, which carries the past sequential information.

Language modelling is one evaluation method to analyse the long-term dependencies of LTM. The Penn treebank and Text8 datasets require longer learning capabilities.  Language modelling requires a clear understanding of the entire text, rather than a window of text. Holding an entire article in order to predict and understand text is easier for the model. LTM through scaling holds all the information passed through the LTM. Therefore, LTM is capable of understanding a clear picture of the entire article. Attention-based memory networks [9] identify the most relevant information and the network predicts based on the information the attention has capture. Attention-based memory networks are capable of handling shorter sequences. It failed to hold long sequence. The attention diverts when given longer sequences. LTM does not focus on memory and holds all past inputs. Unlike attention based networks would forget the most irreverent information which might be relevant later on the sequence, LTM would hold all the information passed through the model.

Table I and Table II compare the LTM with other state-of-the-art models and traditional memory networks. This shows that LTM is capable of handling longer sequences and produces state of the art results. LTM's longer memory plays a crucial role in language modelling tasks. Table III shows that increasing LTM cells would further enhance the results and produce lower perplexity score. LTM has shown to hold longer sequences and be unaffected by vanishing and exploding gradient. 

Similar to LSTM, LTM avoids vanishing or exploding gradient decent using gates. LTM uses gates to enhance the input passed to the network. LTM handles long-term dependencies by the use sigmoid functions to scale the new inputs and carry on the past outputs at the gates. LTM handles long sequences through the scaling. The example of ``\textit {I was born in France. I moved to UK when I was 5 years old ... I speak fluent French}" predicting ``\textit{French}" is attainable since the model holds the entire sequence. Holding the entire sequence in the memory supports the model to predict the last word ``\textit{French}". LTM carries forward the entire sequence allowing the models to use the entirety of the sequence to predict the final word, which holds the most important factor that requires the model to predict the last word. LTM is capable of handling vanishing and exploding gradient as well as handling long-term dependencies.  

\begin{figure}[htbp]
\centerline{\includegraphics[scale=0.35] {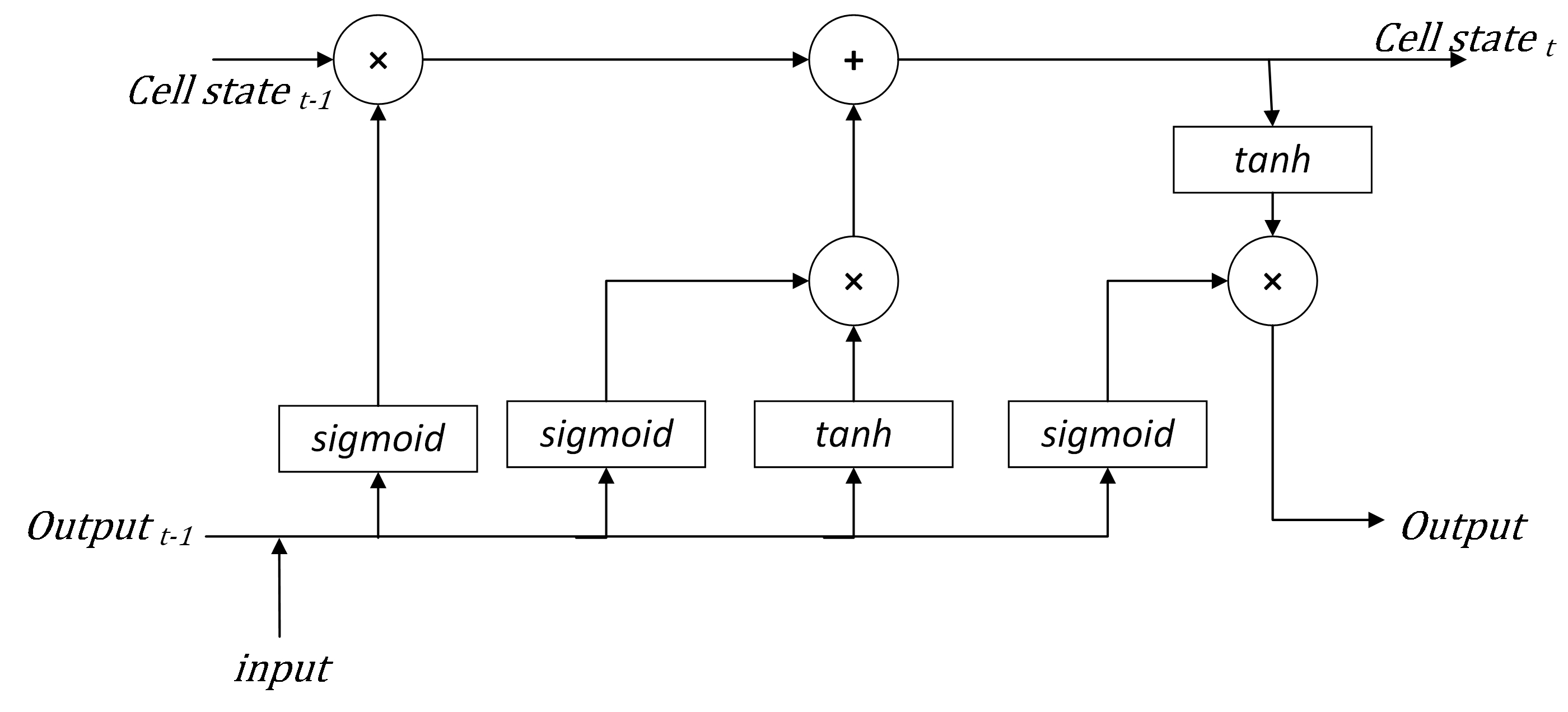}}
\caption{General cell of a Long Short Term Memory network. The figure illustrates a general Long Short Term Memory Network cell taken from the time time stamp $input_{t}$.}
\label{fig2}
\end{figure}
Fig. 3 shows the LSTM cell which holds three gates (forget gate, input gate and output gate). LSTM holds a combination of sigmoid and tanh activation fucntions, while LTM relies only on sigmoid. Comparing Fig. 1 with Fig. 3 indicates the core difference between LSTM and LTM. LTM uses generalization through the sigmoid activation functions hold a longer sequence without forgetting the past information. However, LSTM forgets longer sequences through the forget gates in order to maintain the networks stability. LSTM sacrifies long term dependencies for network stability.

\section{Conclusion}
This paper presents a long-term memory network which is capable of handling long-term dependencies. LTM is capable of handling long sequences without being affected by vanishing or exploding gradient. LTM has shown to outperform traditional LSTM and RNN as well as the memory specific networks in language modelling. LTM was tested on both Penn treebank and Text8 dataset in which LTM has outperformed all state of the art memory networks using minimal hidden units.  Increasing the number of hidden units have shown that the LTM does not get affected by the vanishing and exploding gradient. Adding more hidden unit the LTM has achieved lower perplexity scores and stabilised.

\section*{Acknowledgment}
This work was partially supported by a Murdoch University internal grant on the high-power computer.

\bibliographystyle{IEEEtranN}
\bibliography{references}

\begin{thebibliography}{30}
\providecommand{\natexlab}[1]{#1}
\providecommand{\url}[1]{#1}
\csname url@samestyle\endcsname
\providecommand{\newblock}{\relax}
\providecommand{\bibinfo}[2]{#2}
\providecommand{\BIBentrySTDinterwordspacing}{\spaceskip=0pt\relax}
\providecommand{\BIBentryALTinterwordstretchfactor}{4}
\providecommand{\BIBentryALTinterwordspacing}{\spaceskip=\fontdimen2\font plus
\BIBentryALTinterwordstretchfactor\fontdimen3\font minus
  \fontdimen4\font\relax}
\providecommand{\BIBforeignlanguage}[2]{{%
\expandafter\ifx\csname l@#1\endcsname\relax
\typeout{** WARNING: IEEEtranN.bst: No hyphenation pattern has been}%
\typeout{** loaded for the language `#1'. Using the pattern for}%
\typeout{** the default language instead.}%
\else
\language=\csname l@#1\endcsname
\fi
#2}}
\providecommand{\BIBdecl}{\relax}
\BIBdecl

\bibitem[Nugaliyadde et~al.(2017)Nugaliyadde, Wong, Sohel, and
  Xie]{nugaliyadde2017reinforced}
A.~Nugaliyadde, K.~W. Wong, F.~Sohel, and H.~Xie, ``Reinforced memory network
  for question answering,'' in \emph{International Conference on Neural
  Information Processing}.\hskip 1em plus 0.5em minus 0.4em\relax Springer,
  2017, pp. 482--490.

\bibitem[Bahdanau et~al.(2014)Bahdanau, Cho, and Bengio]{bahdanau2014neural}
D.~Bahdanau, K.~Cho, and Y.~Bengio, ``Neural machine translation by jointly
  learning to align and translate,'' \emph{arXiv preprint arXiv:1409.0473},
  2014.

\bibitem[Yang et~al.(2017)Yang, Chen, Wang, and Xu]{yang2017multi}
Z.~Yang, W.~Chen, F.~Wang, and B.~Xu, ``Multi-sense based neural machine
  translation,'' in \emph{Neural Networks (IJCNN), 2017 International Joint
  Conference on}.\hskip 1em plus 0.5em minus 0.4em\relax IEEE, 2017, pp.
  3491--3497.

\bibitem[Mikolov et~al.(2014)Mikolov, Joulin, Chopra, Mathieu, and
  Ranzato]{mikolov2014learning}
T.~Mikolov, A.~Joulin, S.~Chopra, M.~Mathieu, and M.~Ranzato, ``Learning longer
  memory in recurrent neural networks,'' \emph{arXiv preprint arXiv:1412.7753},
  2014.

\bibitem[Ororbia~II et~al.(2017)Ororbia~II, Mikolov, and
  Reitter]{ororbia2017learning}
A.~G. Ororbia~II, T.~Mikolov, and D.~Reitter, ``Learning simpler language
  models with the differential state framework,'' \emph{Neural computation},
  vol.~29, no.~12, pp. 3327--3352, 2017.

\bibitem[Singh and Lee(2017)]{singh2017temporal}
M.~D. Singh and M.~Lee, ``Temporal hierarchies in multilayer gated recurrent
  neural networks for language models,'' in \emph{Neural Networks (IJCNN), 2017
  International Joint Conference on}.\hskip 1em plus 0.5em minus 0.4em\relax
  IEEE, 2017, pp. 2152--2157.

\bibitem[Hochreiter and Schmidhuber(1997)]{hochreiter1997long}
S.~Hochreiter and J.~Schmidhuber, ``Long short-term memory,'' \emph{Neural
  computation}, vol.~9, no.~8, pp. 1735--1780, 1997.

\bibitem[Weston et~al.(2014)Weston, Chopra, and Bordes]{json2014memnet}
J.~Weston, S.~Chopra, and A.~Bordes, ``Memory networks,'' \emph{arXiv preprint
  arXiv:1410.3916}, 2014.

\bibitem[Bengio et~al.(1994)Bengio, Simard, and Frasconi]{bengio1994learning}
Y.~Bengio, P.~Simard, and P.~Frasconi, ``Learning long-term dependencies with
  gradient descent is difficult,'' \emph{IEEE transactions on neural networks},
  vol.~5, no.~2, pp. 157--166, 1994.

\bibitem[Gers et~al.(1999)Gers, Schmidhuber, and Cummins]{gers1999learning}
F.~A. Gers, J.~Schmidhuber, and F.~Cummins, ``Learning to forget: Continual
  prediction with lstm,'' 1999.

\bibitem[Nugaliyadde et~al.(2019)Nugaliyadde, Wong, Sohel, and
  Xie]{nugaliyadde2019enhancing}
A.~Nugaliyadde, K.~W. Wong, F.~Sohel, and H.~Xie, ``Enhancing semantic word
  representations by embedding deeper word relationships,'' \emph{arXiv
  preprint arXiv:1901.07176}, 2019.

\bibitem[Graves et~al.(2014)Graves, Wayne, and Danihelka]{graves2014neural}
A.~Graves, G.~Wayne, and I.~Danihelka, ``Neural turing machines,'' \emph{arXiv
  preprint arXiv:1410.5401}, 2014.

\bibitem[Sukhbaatar et~al.(2015)Sukhbaatar, Weston, Fergus,
  et~al.]{sukhbaatar2015end}
S.~Sukhbaatar, J.~Weston, R.~Fergus \emph{et~al.}, ``End-to-end memory
  networks,'' in \emph{Advances in neural information processing systems},
  2015, pp. 2440--2448.

\bibitem[Weston et~al.(2015)Weston, Bordes, Chopra, Rush, van Merri{\"e}nboer,
  Joulin, and Mikolov]{weston2015towards}
J.~Weston, A.~Bordes, S.~Chopra, A.~M. Rush, B.~van Merri{\"e}nboer, A.~Joulin,
  and T.~Mikolov, ``Towards ai-complete question answering: A set of
  prerequisite toy tasks,'' \emph{arXiv preprint arXiv:1502.05698}, 2015.

\bibitem[Boukoros et~al.(2017)Boukoros, Nugaliyadde, Marnerides, Vassilakis,
  Koutsakis, and Wong]{boukoros2017modeling}
S.~Boukoros, A.~Nugaliyadde, A.~Marnerides, C.~Vassilakis, P.~Koutsakis, and
  K.~W. Wong, ``Modeling server workloads for campus email traffic using
  recurrent neural networks,'' in \emph{International Conference on Neural
  Information Processing}.\hskip 1em plus 0.5em minus 0.4em\relax Springer,
  2017, pp. 57--66.

\bibitem[Pascanu and Bengio(1986)]{pascanu1986learning}
R.~Pascanu and Y.~Bengio, ``Learning to deal with long-term dependencies,''
  \emph{Neural Computation}, vol.~9, pp. 1735--1780, 1986.

\bibitem[Salehinejad(2016)]{salehinejad2016learning}
H.~Salehinejad, ``Learning over long time lags,'' \emph{arXiv preprint
  arXiv:1602.04335}, 2016.

\bibitem[Pascanu et~al.(2013)Pascanu, Mikolov, and
  Bengio]{pascanu2013difficulty}
R.~Pascanu, T.~Mikolov, and Y.~Bengio, ``On the difficulty of training
  recurrent neural networks,'' in \emph{International Conference on Machine
  Learning}, 2013, pp. 1310--1318.

\bibitem[Hochreiter et~al.(2001)Hochreiter, Bengio, Frasconi, Schmidhuber,
  et~al.]{hochreiter2001gradient}
S.~Hochreiter, Y.~Bengio, P.~Frasconi, J.~Schmidhuber \emph{et~al.}, ``Gradient
  flow in recurrent nets: the difficulty of learning long-term dependencies,''
  2001.

\bibitem[Young et~al.(2018)Young, Hazarika, Poria, and
  Cambria]{young2018recent}
T.~Young, D.~Hazarika, S.~Poria, and E.~Cambria, ``Recent trends in deep
  learning based natural language processing,'' \emph{ieee Computational
  intelligenCe magazine}, vol.~13, no.~3, pp. 55--75, 2018.

\bibitem[Chen et~al.(1998)Chen, Hwang, and Wang]{chen1998rnn}
S.-H. Chen, S.-H. Hwang, and Y.-R. Wang, ``An rnn-based prosodic information
  synthesizer for mandarin text-to-speech,'' \emph{IEEE transactions on speech
  and audio processing}, vol.~6, no.~3, pp. 226--239, 1998.

\bibitem[Mikolov et~al.(2010)Mikolov, Karafi{\'a}t, Burget, {\v{C}}ernock{\`y},
  and Khudanpur]{mikolov2010recurrent}
T.~Mikolov, M.~Karafi{\'a}t, L.~Burget, J.~{\v{C}}ernock{\`y}, and
  S.~Khudanpur, ``Recurrent neural network based language model,'' in
  \emph{Eleventh Annual Conference of the International Speech Communication
  Association}, 2010.

\bibitem[Kumar et~al.(2016)Kumar, Irsoy, Ondruska, Iyyer, Bradbury, Gulrajani,
  Zhong, Paulus, and Socher]{kumar2016ask}
A.~Kumar, O.~Irsoy, P.~Ondruska, M.~Iyyer, J.~Bradbury, I.~Gulrajani, V.~Zhong,
  R.~Paulus, and R.~Socher, ``Ask me anything: Dynamic memory networks for
  natural language processing,'' in \emph{International Conference on Machine
  Learning}, 2016, pp. 1378--1387.

\bibitem[LeCun et~al.(2015)LeCun, Bengio, and Hinton]{lecun2015deep}
Y.~LeCun, Y.~Bengio, and G.~Hinton, ``Deep learning,'' \emph{nature}, vol. 521,
  no. 7553, p. 436, 2015.

\bibitem[Sak et~al.(2014)Sak, Senior, and Beaufays]{sak2014long}
H.~Sak, A.~Senior, and F.~Beaufays, ``Long short-term memory recurrent neural
  network architectures for large scale acoustic modeling,'' in \emph{Fifteenth
  annual conference of the international speech communication association},
  2014.

\bibitem[Chorowski et~al.(2015)Chorowski, Bahdanau, Serdyuk, Cho, and
  Bengio]{chorowski2015attention}
J.~K. Chorowski, D.~Bahdanau, D.~Serdyuk, K.~Cho, and Y.~Bengio,
  ``Attention-based models for speech recognition,'' in \emph{Advances in
  neural information processing systems}, 2015, pp. 577--585.

\bibitem[Cambria and White(2014)]{cambria2014jumping}
E.~Cambria and B.~White, ``Jumping nlp curves: A review of natural language
  processing research,'' \emph{IEEE Computational intelligence magazine},
  vol.~9, no.~2, pp. 48--57, 2014.

\bibitem[Taylor et~al.(2003)Taylor, Marcus, and Santorini]{taylor2003penn}
A.~Taylor, M.~Marcus, and B.~Santorini, ``The penn treebank: an overview,'' in
  \emph{Treebanks}.\hskip 1em plus 0.5em minus 0.4em\relax Springer, 2003, pp.
  5--22.

\bibitem[Mikolov et~al.(2011)Mikolov, Kombrink, Burget, {\v{C}}ernock{\`y}, and
  Khudanpur]{mikolov2011extensions}
T.~Mikolov, S.~Kombrink, L.~Burget, J.~{\v{C}}ernock{\`y}, and S.~Khudanpur,
  ``Extensions of recurrent neural network language model,'' in
  \emph{Acoustics, Speech and Signal Processing (ICASSP), 2011 IEEE
  International Conference on Acoustics, Speech and Signal Processing}.\hskip
  1em plus 0.5em minus 0.4em\relax IEEE, 2011, pp. 5528--5531.

\bibitem[Xie et~al.(2017)Xie, Wang, Li, L{\'e}vy, Nie, Jurafsky, and
  Ng]{xie2017data}
Z.~Xie, S.~I. Wang, J.~Li, D.~L{\'e}vy, A.~Nie, D.~Jurafsky, and A.~Y. Ng,
  ``Data noising as smoothing in neural network language models,'' \emph{arXiv
  preprint arXiv:1703.02573}, 2017.

\end{thebibliography}

\end{document}